\documentclass[runningheads]{llncs}
\usepackage[T1]{fontenc}
\usepackage{array}

\newcolumntype{P}[1]{>{\centering\arraybackslash}p{#1}}
\usepackage{graphicx}
\usepackage{multirow}
\usepackage{soul}
\newcommand{\repeatthanks}{\textsuperscript{\thefootnote}}

\begin{document}

\author{Raheleh Salehi\inst{1,2}\thanks{equal contribution}\and Ario Sadafi\inst{1,3}\repeatthanks\and Armin Gruber\inst{1}\and Peter Lienemann\inst{1}\and\\ Nassir Navab \inst{3,4}\and Shadi Albarqouni\inst{5,6,7} \and Carsten Marr\inst{1}\thanks{carsten.marr@helmholtz-muenchen.de}
}


\institute{Institute of AI for Health, Helmholtz Munich – German Research Center for Environmental Health, Neuherberg, Germany\and
Computer Engineering, Politecnico Di Torino, Italy\and Computer Aided Medical Procedures (CAMP), Technical University of Munich, Germany\and Computer Aided Medical Procedures, Johns Hopkins University, USA\and
Clinic for Interventional and Diagnostic Radiology, University Hospital Bonn, Bonn, Germany\and
Faculty of Informatics, Technical University Munich, Munich, Germany
\and 
Helmholtz AI, Helmholtz Munich – German Research Center for Environmental Health, Neuherberg, Germany
}
\title{Unsupervised Cross-Domain Feature Extraction for Single Blood Cell Image Classification}
\authorrunning{R. Salehi et al.}
\titlerunning{AE-CFE: Unsupervised Cross-Domain Feature Extraction}
%
%
%

%

\maketitle              

\begin{abstract}
Diagnosing hematological malignancies requires identification and classification of white blood cells in peripheral blood smears. Domain shifts caused by different lab procedures, staining, illumination, and microscope settings hamper the re-usability of recently developed machine learning methods on data collected from different sites. 
Here, we propose a cross-domain adapted autoencoder to extract features in an unsupervised manner on three different datasets of single white blood cells scanned from peripheral blood smears. The autoencoder is based on an R-CNN architecture allowing it to focus on the relevant white blood cell and eliminate artifacts in the image.
To evaluate the quality of the extracted features we use a simple random forest to classify single cells. We show that thanks to the rich features extracted by the autoencoder trained on only one of the datasets, the random forest classifier performs satisfactorily on the unseen datasets, and outperforms published oracle networks in the cross-domain task. 
Our results suggest the possibility of employing this unsupervised approach in more complicated diagnosis and prognosis tasks without the need to add expensive expert labels to unseen data.


\end{abstract}

\goodbreak
\section{Introduction}
Hematopoietic malignancies such as leukemkia are among the deadliest diseases with limited therapeutic options. Cytomorphological evaluation of white blood cells under the microscopic in blood or bone marrow smears is key for proper diagnosis. So far, this morphological analysis has not been automated and is still performed manually by trained experts under the microscope. 
Recent works demonstrate however the potential in automation of this task. 
Matek et al. \cite{matek2019human} have proposed a highly accurate approach based on ResNext \cite{xie2017aggregated} architecture for recognition of white blood cells in blood smears of acute myeloid leukemia patients. In another work \cite{matek2021highly} we have developed a CNN-based classification method for cell morphologies in bone marrow smears. Boldu et al. \cite{boldu2019automatic} have suggested a machine learning approach for diagnosis of acute leukemia by recognition of blast cells in blood smear images. Acevedo et al. \cite{acevedo2021new} suggest a predictive model for automatic recognition of patients suffering from myelodysplastic syndrome, a pre-form of acute myeloid leukemia. 

All of these studies have used data provided from a single site. However, many factors in laboratory procedures can affect the data and introduce a domain shift: Different illuminations, microscope settings, camera resolutions, and staining protocols are only some of the parameters differing between laboratories and hospitals. These changes can affect model performance considerably and render established approaches ineffective, requiring re-annotation and re-training of models.

Exposing the optimization to domain shifts can be a solution to align different domains in real-world data. A learning paradigm with dedicated losses is a common way to tackle this problem and has been already applied in many approaches \cite{tolstikhin2016minimax}. For instance, Duo et al. \cite{dou2019domain} propose to learn semantic feature spaces by incorporating global and local constraints in a supervised method, while Chen et al. \cite{chen2020unsupervised} have developed a method for unsupervised domain adaptation by conducting synergistic alignment of both image and features and applied it to medical image segmentation in bidirectional cross-modality adaptation between MRI and CT. 

Here, we present an AutoEncoder-based Cell Feature Extractor (AE-CFE), a simple and economic approach for robust feature extraction of single cells. Our method is based on instance features extracted by a Mask R-CNN \cite{he2017mask} architecture that is analyzed by an autoencoder to obtain features of single white blood cells in digitized blood smears. Since the data is coming from different sites, we are introducing a domain adaptation loss to reduce domain shifts. Our method is the first unsupervised two-staged autoencoder approach for cross-domain feature extraction based on instance features of a Mask R-CNN. It outperforms  published supervised methods in unseen white blood cell datasets and can thus contribute to the establishment of robust decision support algorithms for diagnosing hematopoietic malignancies. 
We made our implementation publicly available at 
\url{https://github.com/marrlab/AE-CFE}

\section{Methodology}

\begin{figure}[t]
\centering
\label{figoverview}
\includegraphics[width=0.9\textwidth,page=1,trim=0cm 4cm 0cm 0cm,clip]{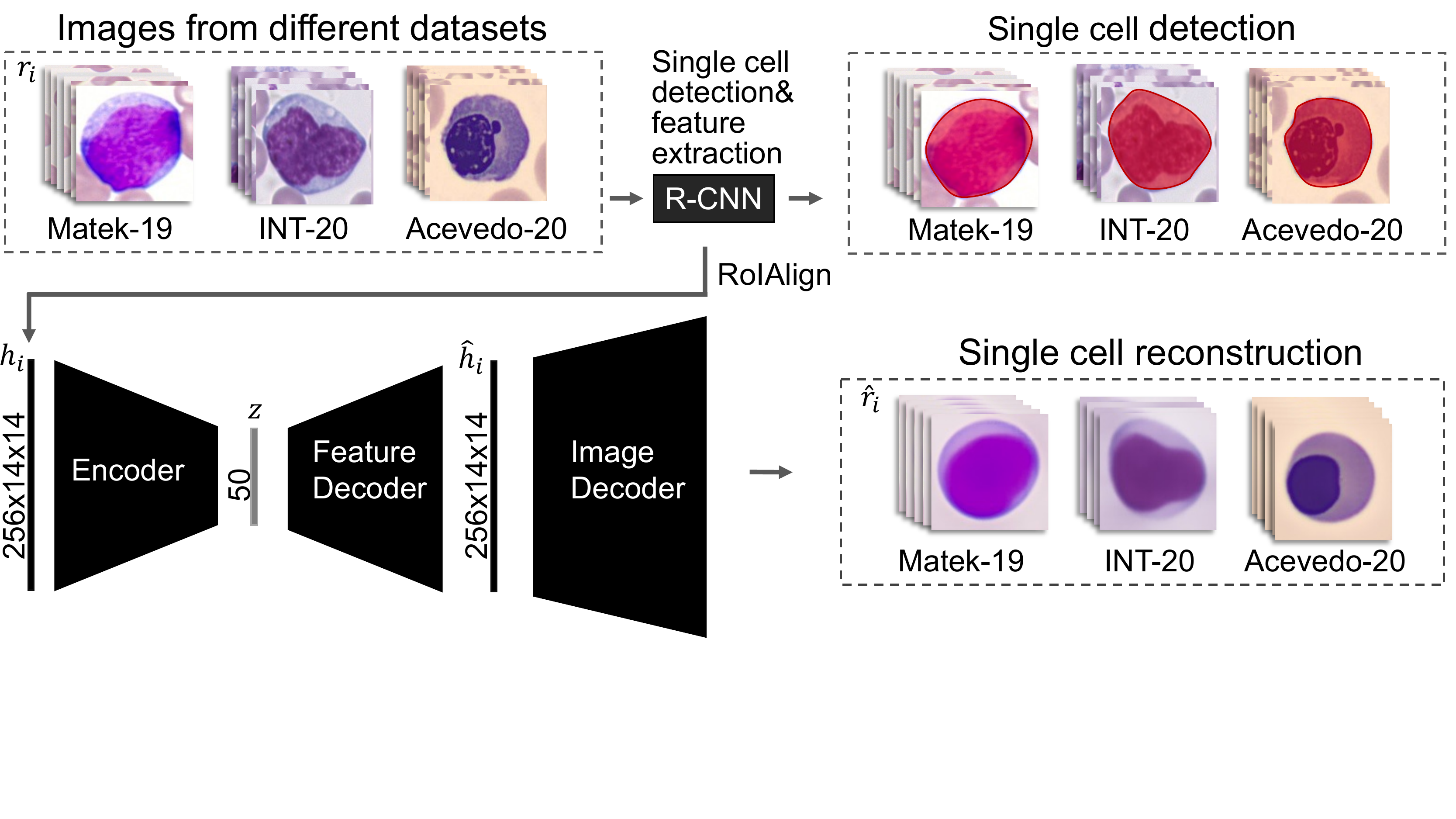}
\caption{Overview of the proposed AE-CFE method. A Mask R-CNN detects single cells in images and relevant instance features of the region of interest are extracted. The autoencoder uses the instance features as input and tries to reconstruct (i) instance features and (ii) single cell images. Since features are white blood cell specific, the autoencoder is able to only reconstruct white blood cells and artefacts such as red blood cells are discarded.}
\end{figure}

Our unsupervised feature extraction approach starts with a Mask R-CNN \cite{he2017mask} model trained to detect single white blood cells in scanned patient's blood smears. For every detected cell instance-specific features extracted are used for training an autoencoder. This compresses the input to a latent space representation, while a two-staged decoder tries to reconstruct (i) the encoded features and (ii) the single cell images. 

Mask R-CNN is commonly used for instance segmentation. The architecture is based on an underlying feature extractor based on a ResNet-101-FPN \cite{he2016deep,lin2017feature} backbone. It has two stages: (i) A region proposal network (RPN) suggests candidate bounding boxes all over the input image and (ii) different heads of the architecture perform classification, bounding box regression, and segmentation locally only on the region of interest (RoI) based on the features that are extracted for every instance with RoIAlign. 

More formally, having an image $I_i$ from dataset $D_k$, 
\begin{equation}
    r_{i,j}, h_{i,j} = f_{\mathrm{R-CNN}} (I_i) : \forall I_i \in D_k \ , 
\end{equation}
where $r_{i,j}$ is the $j^{\mathrm{th}}$ single cell image cropped out and, $h_{i,j}$ is its corresponding features in $i^{\mathrm{th}}$ image of the dataset. For simplicity we assume for now there is only one white blood cell in every image and refer to it with $r_i$ and $h_i$ in the rest of this section.  

Our desired feature extraction method can be formulated as 
\begin{equation}
    z = f_{\mathrm{enc}}(h_i; \theta) \ ,
\end{equation}
where $z$ is the robust, cross-domain feature vector we get at the bottleneck of the autoencoder and $\theta$ are the parameters learned during the training. 

The autoencoder consists of three modules: (i) encoder, (ii) feature decoder, and (iii) image decoder. All three parts are trained by optimizing

\begin{equation}
\label{eqloss}
    \mathcal{L}(\theta,\gamma,\psi) =
    \frac{1}{N} \sum_{i=1}^N(\hat{h}_i - h_i)^2 +1 - \mathrm{SSIM}(\hat{r}_i, r_i) +  \beta \mathcal{L}_{\mathrm{DA} \ ,}
\end{equation}
where $\hat{h}_i = f^{\mathrm{feat}}_{\mathrm{dec}} (z ; \gamma)$ is the reconstructed feature vector, $\hat{r}_i = f^{\mathrm{img}}_{\mathrm{dec}} (\hat{h}_i; \psi)$ is the reconstructed image based on the feature reconstruction, $\gamma$ and $\psi$ are model parameters, and $N$ is number of white blood cells in the dataset. $\mathcal{L}_{\mathrm{DA}}$ is the domain adaptation loss introduced in section \ref{DAloss} regulated by constant coefficient $\beta$. 
We use the structural similarity index measure (SSIM) \cite{wang2004image} to measure the similarity of the reconstructed image $x$ with the original single cell image $y$ detected by the Mask R-CNN, defined as
\begin{equation}
    \mathrm{SSIM}(x,y) = \frac{(2 \mu_x\mu_y + c_1)(2\sigma_{xy} + c_2)}{(\mu_x^2 + \mu_y^2 + c_1)(\sigma_x^2 + \sigma_y^2 + c_2)}
\end{equation}
where $\mu$ and $\sigma$ are the mean and variance of the images and $c_1$, $c_2$ are small constants for numerical stability. 

Group normalization (GN) \cite{wu2018group} is applied after each layer in the encoder part as an alternative to batch normalization that is not dependent on the batch size. It divides the channels into groups and normalizes the groups with independent group specific mean and variance. In our experiments GN was effective in image generalization.

\begin{figure}[t]
\centering

\includegraphics[width=0.8\textwidth,page=2,trim=0cm 0.5cm 0cm 0cm,clip]{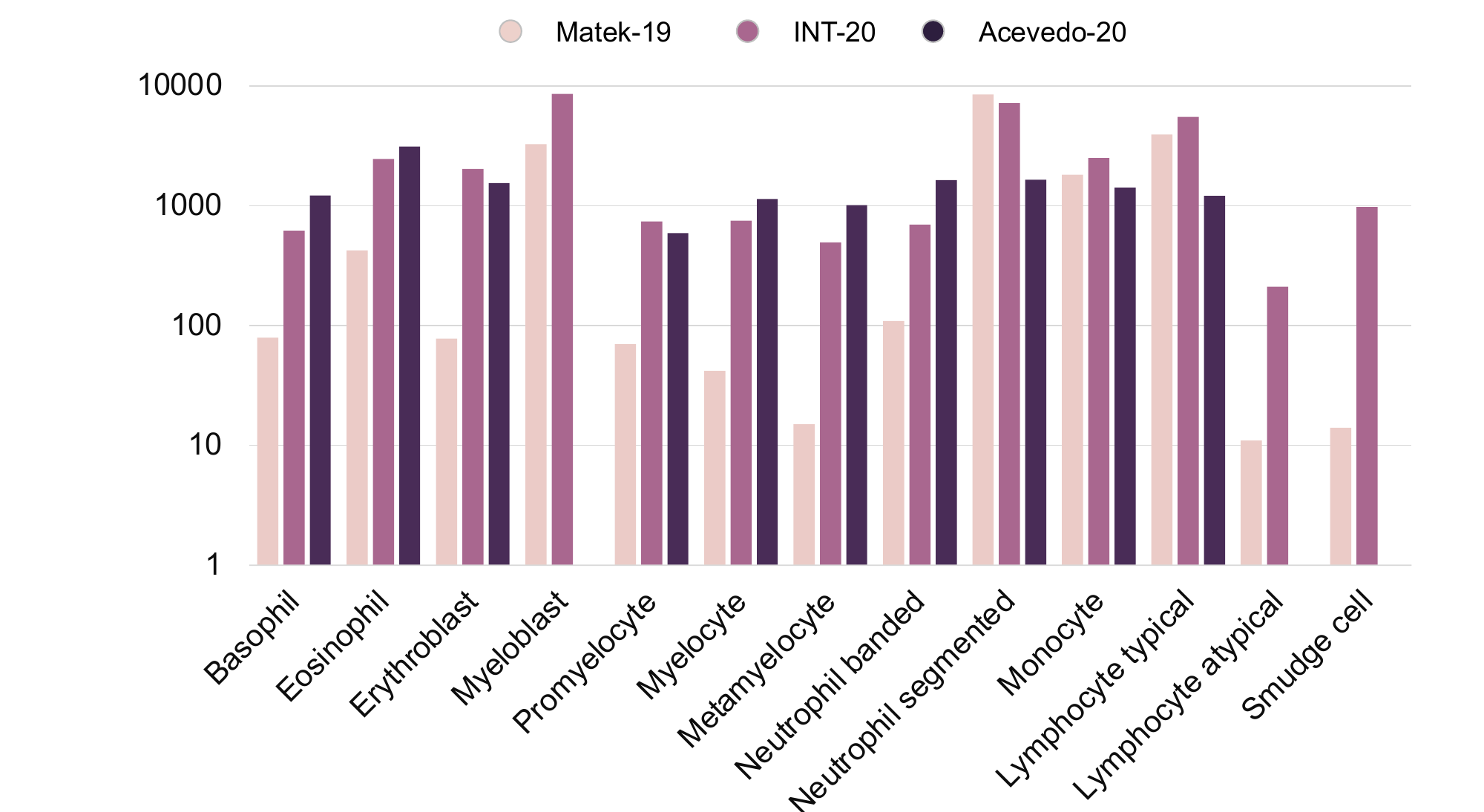}
\caption{Number of samples in each of the 13 classes for the three datasets used in our study.}
\label{figdataset}
\end{figure}

\subsection{Domain adaptation} \label{DAloss}
When images come from different sites, the latent space can be dominated by a domain shift (see Fig. \ref{figlat}). Domain adaptation with group normalization and distribution-based maximum mean discrepancy has been shown to align the latent space representation of different datasets \cite{tzeng2017adversarial}. We use it to adapt the three datasets, which differ in resolution, size, and color.\\
With $\mathcal{D} = \{D_1, \ldots D_K\}$ being the datasets we are training on, and a mean matrix $\mu_k$ and $s_k$ as softmax of covariance matrix of the embedded features of dataset $D_k$, our loss is defined by

\begin{equation}
    \mathcal{L}_{\mathrm{DA}} =\sum_{k=1}^{K}\{ {\mathrm{MSE}(\mu_k, \mu_0)} + \frac{1}{2}[D_{KL} (s_0||s_k) + D_{KL} (s_k||s_0) ]\} \ ,
\end{equation}
where we calculate mean squared error on the mean matrices and a symmetrized Kullback-Leibler (KL) divergence of the covariance matrices for all datasets to bring them closer to the anchor dataset $D_0$. Any other symmetric divergences, or cosine similarity between the eigenvectors of the covariance matrices would work for this optimization.

\section{Evaluation}

\subsection{Datasets}
We are using three different datasets to evaluate our method: 

The \textbf{Matek-19} dataset consists of over 18,000 annotated white blood cells from 100 acute myeloid leukaemia patients and 100 patients exhibiting no morphological features from the laboratory of leukemia diagnostics at Munich University Hospital between 2014 and 2017. It is publicly available \cite{matek2019human} and there are 15 classes in the dataset. Image dimensions are $400\times400$ pixels or approximately $29\times29$ micrometers.

The \textbf{INT-20} in-house dataset has around 42,000 images coming from 18 different classes. Images are $288\times288$ in pixels or $25\times25$ micrometers.

The \textbf{Acevedo-20} dataset consists of over 17,000 images of individual normal cells acquired in the core laboratory at the Hospital Clinic of Barcelona published by Acevedo et al. \cite{acevedo2020dataset}. There are 8 classes in the dataset and images are $360\times363$ pixels or $36\times36.3$ micrometers.

Since class definitions of the three datasets are different, we asked a medical expert to categorize different labels into 13 commonly defined classes consisting of: basophil, eosinophil, erythroblast, myeloblast, promyelocyte, myelocyte, metamyelocyte, neutrophil banded, neutrophil segmented, monocyte, lymphocyte typical, lymphocyte atypical, and smudge cells. Figure \ref{figdataset} shows sample distribution between these 13 classes for different datasets.

\subsection{Implementation details}
\begin{figure}[t]
\centering
\includegraphics[width=1\textwidth,page=3,trim=2cm 9cm 1cm 1cm,clip]{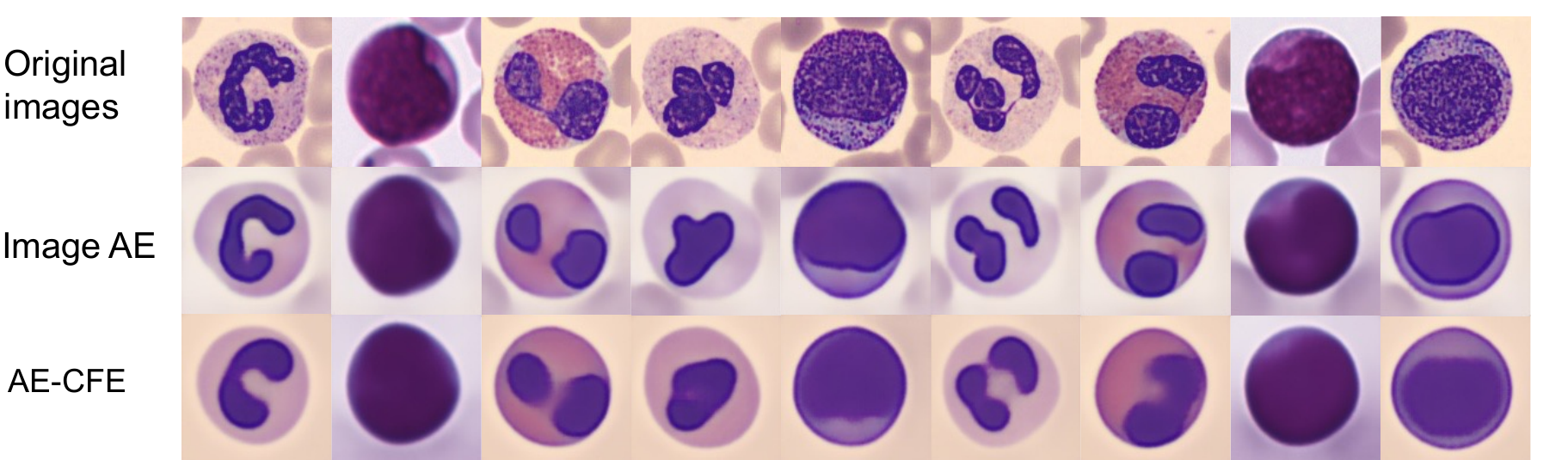}
\caption{We are comparing the reconstruction of AE-CFE with an image based autoencoder (Image AE). Red blood cells and artifacts surrounding the white blood cells are eliminated with Mask R-CNN feature extraction.}
\label{figmrcnn}
\end{figure}
\subsubsection{Architecture}
The autoencoder is a fully convolutional network. The encoder consists of 6 layers, the feature decoder has 3 layers, and the image decoder has 5 layers. All intermediate layers have ReLU activation functions, and outputs of encoder and feature decoder are regulated by a tanh activation function while the image decoder has a sigmoid activation function on the output. To extract richest and least sparse features, we decided to use 50 as the bottleneck size which is the smallest possible.

\subsubsection{Training} 
We performed stratified train and test splits on all datasets keeping 20\% of the data for a holdout test set. 
Training was carried out using an Adam optimizer for 150 epochs with a learning rate of $0.001$ on three NVIDIA A100-SXM4-40GB GPUs with a total batch size of 1500 (500 on each). The constant $\beta$ in equation \ref{eqloss} was set to $5$. 

\subsubsection{Random Forest}
We used random forest implementation of the scikit-learn library \cite{sklearn_api} for all of the experiments. Number of estimators was set to 200 and maximum tree depth to 16.

\subsection{Single cell detection by Mask R-CNN}

\begin{figure}[t]
\centering
\includegraphics[width=0.8\textwidth,page=4,trim=0cm 8.5cm 9cm 0cm,clip]{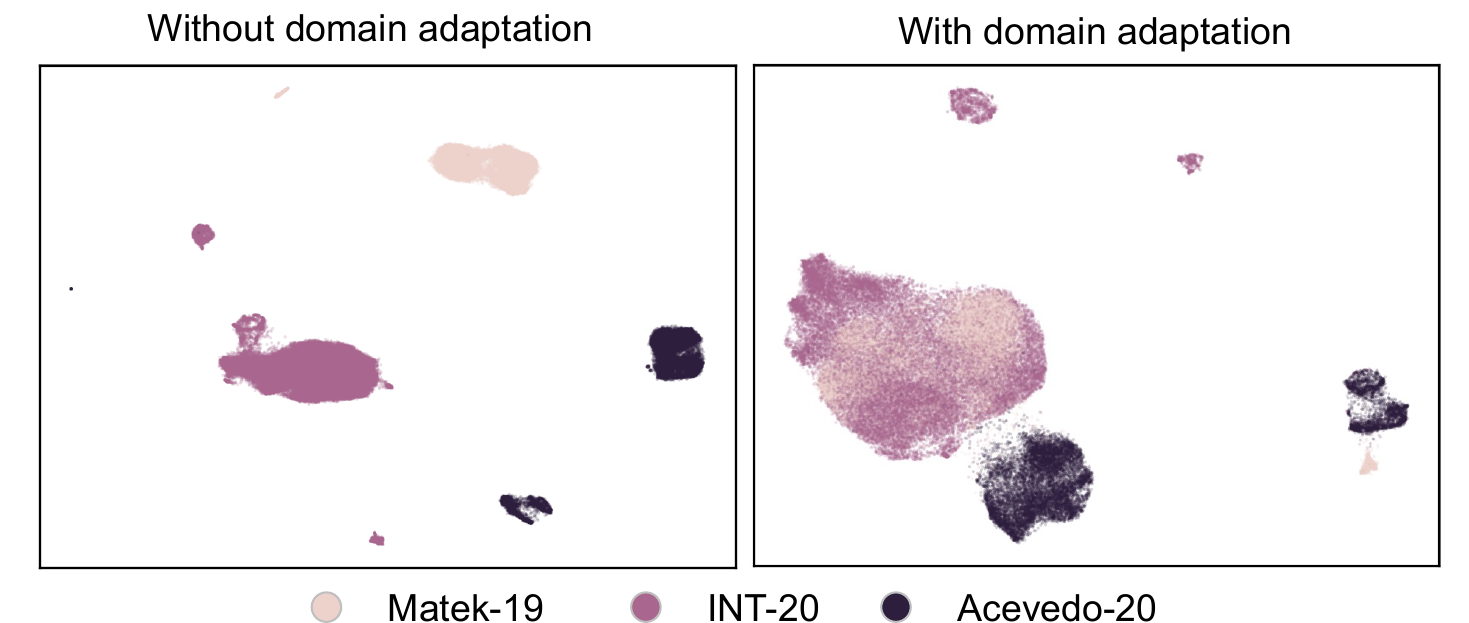}
\caption{UMAP embedding of AE-CFE with and without the domain adaptation loss. A more uniformly distributed latent representation is achieved after using the domain adaptation loss.}
\label{figlat}
\end{figure}

The Mask R-CNN helps extracting the instance features and eliminate artefacts. To verify this observation, we trained another autoencoder with a similar number of layers in the decoder section on single white blood cell images as a baseline. Figure \ref{figmrcnn} shows example reconstructions from both methods. The trained autoencoder is affected by the noise from surrounding red blood cells way more than AE-CFE. 

The Mask R-CNN model is trained on a small separate dataset of around 1500 images annotated for instance segmentation. The annotation of this dataset does not require any expertise, as white blood cell shapes are annotated with no class information. The Mask R-CNN was trained for 26 epoches reaching a mAP of 0.89. 
Analysing a pool of 77,363 images coming from three datasets, 65,693 cells were successfully detected (85\%).

\subsection{Evaluation}
To quantitatively compare the quality of the extracted features by AE-CFE, we train a random forest (RF) model on the extracted features trying to classify single white blood cells into one of the 13 defined classes. 
\begin{table}

\caption{Comparing the accuracy percentage of a random forest method trained on our proposed AE-CFE approach with 4 other feature extraction methods as baselines: ResNet101 trained on ImageNet, features extracted with Mask R-CNN, an autoencoder trained on instance feature vectors, and an adversarial autoencoder trained on instance feature vector. Mean and standard deviation of accuracy is reported from 5 training runs.}
\begin{tabular}{l|l||c|c|c|c|c}
\label{tblBaselines}
 Trained on & Tested on & Reset-RF & R-CNN-RF & AE-RF & AAE-DA & AE-CFE \\ \hline 
\multirow{3}{*}{ Matek-19 } & Matek-19 &  62.5±1.8 &  60.5±0.8 &  86.0±0.04 & \textbf{ 87.5±0.8 } & 83.7±0.5 \\ \cline{2-7}
& INT-20 & 0 & 0 & 46.8±0.2 & 31.4±0.3 & \textbf{ 48.4±0.2 } \\ \cline{2-7} 
& Acevedo-20 & 0 & 0 & 20.1±0.1 & 18.6±0.4 & \textbf{ 21.9±0.4 } \\ \cline{1-7}
\multirow{3}{*}{ INT-20 } & Matek-19 & 0 & 0 & 47.2±3.4 & 63.9±0.2 & \textbf{73.2±0.1 }\\ \cline{2-7} & INT-20 & 45.2±1.1 & 46.0±0.4 & \textbf{69.1±0.4} & 66.8±0.4 & 65.6±0.5 \\ \cline{2-7} & Acevedo-20 & 0 & 0 & 4.6±0.6 & 17.7±0.7 & \textbf{31.8±0.4 } \\ \cline{1-7}  
\multirow{3}{*}{ Acevedo-20 } & Matek-19 &  0 & 0 & 39.5±1.4 & 39.4±0.6 & \textbf{45.1±0.5 } \\ \cline{2-7} 
& INT-20 & 0 & 0 & 9.7±0.3 & 17.7±0.7 & \textbf{ 21.0±0.5 } \\ \cline{2-7} 
& Acevedo-20 & 37.1±0.8 & 35.9±1.1 & \textbf{67.2±0.7} & 64.3±0.1 & 65.2±0.5
\end{tabular}
\begin{tabular}{l|l|c|c|c|c}
\end{tabular}
\end{table}

In all experiments, the RF is trained on one dataset and tested on the test set of all three datasets. We defined four baselines for our proposed method: (i) ResNet-RF: random forest classification of the features extracted with a ResNet101 \cite{he2016deep} architecture trained on ImageNet dataset \cite{imagenet_cvpr09} (ii) R-CNN-RF: random forest classification of the instance features extracted with our trained Mask R-CNN architecture (iii) AE-RF: random forest classification of features extracted by a similar autoencoder trained on all datasets with no domain adaptation. (iv) AAE-DA: trained  Adversarial domain adaption on features extracted by a similar autoencoder.

In Table \ref{tblBaselines} we compare the accuracy of random forest classification of our method with the baselines and report mean and standard deviation of accuracy for 5 runs. For two of the baselines (ResNet-RF \& R-CNN-RF) cross-domain evaluations were inaccurate (accuracy close to zero) and random forest was unable to classify any sample correctly.

Next, we compare our method with oracle methods specifically trained for classifying the datasets. Matek et al. \cite{matek2019human} have published their ResNext architecture and trained model weights.
We trained a similar ResNext model on each of the datasets.
In Table \ref{tblorac} and supplementary material, we compare these oracle methods with RF trained on features extracted with our method. 
We find that both of the oracles are failing on the unseen dataset while a random forest trained on our unsupervised AE-CFE features is performing by far better.

\begin{table}[h]
\caption{Comparing the accuracy percentage of a random forest method trained on our proposed AE-CFE feature extraction approach with 2 other oracle methods specifically trained for each of the datasets. Matek et al.'s published method trained on their dataset, and two ResNext models trained on each of the datasets with a random forest classifying features of our proposed cross-domain autoencoder. Mean and standard deviation of accuracy is reported over 5 runs.}
\begin{center}
\begin{tabular}{p{2cm}|p{2cm}|P{2.6cm}|P{2.3cm}|P{2.6cm}}
Trained on & Tested on & ResNext & Matek et al. & AE-CFE \\ \hline
\multirow{3}{*}{Matek-19} & Matek-19 & - & \textbf{96.1} & 83.7±0.5 \\ \cline{2-5} 
 & INT-20 & - & 29.5 & \textbf{48.4±0.2} \\ \cline{2-5} 
 & Acevedo-20 & - & 8.1 & \textbf{21.9±0.4} \\ \hline
\multirow{3}{*}{INT-20} & Matek-19 & 49.0±6.3 & - & \textbf{73.2±0.1} \\ \cline{2-5} 
 & INT-20 & \textbf{88.7±1.5} & - & 65.6±0.5 \\ \cline{2-5} 
 & Acevedo-20 & 16.9±1.6 & - & \textbf{31.8±0.4} \\ \hline
\multirow{3}{*}{Acevedo-20} & Matek-19 & 7.3±3.1 & - & \textbf{45.1±0.5} \\ \cline{2-5} 
 & INT-20 & 8.1±1.4 & - & \textbf{21.0±0.5} \\ \cline{2-5} 
 & Acevedo-20 & \textbf{85.7±2.4} & - & 65.2±0.5
\label{tblorac}
\end{tabular}
\end{center}
\end{table}

Finally, we are comparing the UMAP \cite{mcinnes2018umap} embeddings of all feature vectors of the white blood cells from the three datasets with and without our domain adaptation loss. Figure \ref{figlat} shows that with domain adaptation not only the RF classification results improve but also a more uniform latent distribution is achieved, supporting the results.

\section{Discussion \& Conclusion}
Artefacts in single cell images can greatly affect the performance of a model by falsely overfitting on irrelevant features. For example, the surrounding red blood cells and thus the number of red pixels in images can mislead the model into categorizing samples based on anemic features (i.e. the density of red blood cells) rather than the cytomorphological white blood cell properties. This makes Mask R-CNN an essential element in our design, forcing the algorithm to focus on the instance features cropped out in the region of interest rather than the whole image or features in the background. 
But what if cells in unseen data are considerably different? The fact that the training dataset for Mask R-CNN was coming from only one of the three datasets used in our study, 85\% detection rate for single cells in unseen data is surprisingly high, and obviously good enough for demonstrating the multi-domain applicability of our approach. However, annotation of single white blood cells in images from different datasets to train a better Mask R-CNN is cheap, convenient, and fast and can improve our results even further.

For classification, the random forest model trained on features extracted by our approach is not performing as good as oracle models on the source domains, but its performance in cross-domain scenarios is by far superior. 
This is partly due to our domain adaptation loss that forces the latent representations from different datasets to be as close as possible to each other. The small feature vectors of only 50 dimensions with minimum sparsity allow usage of these features in many different applications. 

Using features from cell nuclei additionally, including the AE-CFE approach in decision support algorithms, and testing our method in continuous training scenarios where datasets are added one by one are just some of the exciting directions we plan to follow in the future works.  
Our promising results support the quality of cross-domain cell features extracted by AE-CFE and allow expansion of the developed approaches on new data collected from new sites and hospitals. 

\section*{Acknowledgments}
C.M. has received funding from the European Research Council (ERC) under the European Union’s Horizon 2020 research and innovation programme (Grant agreement No. 866411)

\bibliographystyle{splncs04}
\bibliography{article}

\end{document}


\author{Raheleh Salehi\inst{1,2}\thanks{equal contribution}\and Ario Sadafi\inst{1,3}\repeatthanks\and Armin Gruber\inst{1}\and Peter Lienemann\inst{1}\and\\ Nassir Navab \inst{3,4}\and Shadi Albarqouni\inst{5,6,7} \and Carsten Marr\inst{1}\thanks{carsten.marr@helmholtz-muenchen.de}
}


\institute{Institute of AI for Health, Helmholtz Munich – German Research Center for Environmental Health, Neuherberg, Germany\and
Computer Engineering, Politecnico Di Torino, Italy\and Computer Aided Medical Procedures (CAMP), Technical University of Munich, Germany\and Computer Aided Medical Procedures, Johns Hopkins University, USA\and
Clinic for Interventional and Diagnostic Radiology, University Hospital Bonn, Bonn, Germany\and
Faculty of Informatics, Technical University Munich, Munich, Germany
\and 
Helmholtz AI, Helmholtz Munich – German Research Center for Environmental Health, Neuherberg, Germany
}
\title{Unsupervised Cross-Domain Feature Extraction for Single Blood Cell Image Classification}
\authorrunning{R. Salehi et al.}
\titlerunning{Unsupervised Cross-Domain Feature Extraction}
%
%
%

%
\maketitle              

\section*{Supplementary material}

\begin{table}[h]
\caption{F1 score of a random forest classifier trained on our proposed AE-CFE feature extraction approach with 2 other oracle methods specifically trained for each of the datasets. Matek et al.'s published method and two ResNext models trained on each of the datasets with a random forest classifying features of our proposed cross-domain autoencoder. Mean and standard deviation of weighted F1 score is reported over 5 runs.}
\begin{tabular}{p{2cm}|p{2cm}|P{2.6cm}|P{2.3cm}|P{2.6cm}}
Trained on & Tested on & ResNext & Matek et al. & AE-CFE \\ \hline
\multirow{3}{*}{Matek-19} & Matek-19 & - & \textbf{0.964} & 0.813 ± 0.004 \\ \cline{2-5} 
 & INT-20 & - & 0.301 & \textbf{0.415 ± 0.002} \\ \cline{2-5} 
 & Acevedo-20 & - & 0.034 & \textbf{0.130 ± 0.007} \\ \hline
\multirow{3}{*}{INT-20} & Matek-19 & 0.456 ± 0.024 & - & \textbf{0.700 ± 0.003} \\ \cline{2-5} 
 & INT-20 & \textbf{0.885 ± 0.013} & - & 0.605 ± 0.005 \\ \cline{2-5} 
 & Acevedo-20 & 0.129 ± 0.026 & - & \textbf{0.280 ± 0.003} \\ \hline
\multirow{3}{*}{Acevedo-20} & Matek-19 & 0.078 ± 0.041 & - & \textbf{0.475 ± 0.004} \\ \cline{2-5} 
 & INT-20 & 0.047 ± 0.006 & - & \textbf{0.193 ± 0.011} \\ \cline{2-5} 
 & Acevedo-20 & \textbf{0.854 ± 0.033} & - & 0.632 ± 0.006
\end{tabular}
\end{table}